\documentclass[11pt,a4paper]{article}
\usepackage[hyperref]{acl2019}
\usepackage{times}
\usepackage{latexsym}
\usepackage{graphicx}
\usepackage{multirow}
\usepackage{multicol}
\usepackage{arydshln}
\usepackage{amsmath}
\usepackage{bm}
\usepackage{color}
\usepackage{amssymb}
\usepackage{url}
\usepackage{footmisc}
\usepackage[inference]{semantic}

\usepackage{url}

\aclfinalcopy 
\def\aclpaperid{394} 

\title{Automatic Generation of High Quality CCGbanks \\ for Parser Domain Adaptation}

\author{Masashi Yoshikawa$^1$\\ {\tt yoshikawa.masashi.yh8@}\\{\tt is.naist.jp}\\
        \And
        \hspace{1.5cm}Hiroshi Noji$^2$ \\\hspace{1.5cm} {\tt hiroshi.noji@aist.go.jp}\\
        \AND
        Koji Mineshima$^3$\\{\tt mineshima.koji@ocha.ac.jp}\\
        \And
        \hspace{1.5cm}Daisuke Bekki$^3$\\\hspace{1.5cm}{\tt bekki@is.ocha.ac.jp}\\
        \AND
       $^1${\rm Nara Institute of Science and Technology, Nara, Japan} \\
       $^2${\rm Artificial Intelligence Research Center, AIST, Tokyo, Japan} \\
       $^3${\rm Ochanomizu University, Tokyo, Japan}
}

\date{}

\begin{document}
\maketitle
\begin{abstract}
   We propose a new domain adaptation method for Combinatory Categorial Grammar (CCG) parsing, based on the idea of automatic generation of CCG corpora exploiting cheaper resources of dependency trees.
   Our solution is conceptually simple, and not relying on a specific parser architecture, making it applicable to the current best-performing parsers.
   We conduct extensive parsing experiments with detailed discussion; on top of existing benchmark datasets on (1) biomedical texts and (2) question sentences, we create experimental datasets of (3) speech conversation and (4) math problems. When applied to the proposed method, an off-the-shelf CCG parser shows significant performance gains, improving from 90.7\% to 96.6\% on speech conversation, and from 88.5\% to 96.8\% on math problems.
\end{abstract}

\section{Introduction}
The recent advancement of Combinatory Categorial Grammar (CCG; \citet{steedman:syntactic_process}) parsing~\cite{lee-lewis-zettlemoyer:2016:EMNLP2016,depccg}, combined with formal semantics, has enabled high-performing natural language inference systems~\cite{abzianidze:2017:EMNLP2017Demos,martinezgomez-EtAl:2017:EACLlong}.
We are interested in transferring the success to a range of applications, e.g., inference systems on scientific papers and speech conversation.

To achieve the goal, it is urgent to enhance the CCG parsing accuracy on new domains, i.e., solving a notorious problem of {\it domain adaptation} of a statistical parser, which has long been addressed in the literature.
Especially in CCG parsing, prior work~\cite{D08-1050,lewis-lee-zettlemoyer:2016:N16-1} has taken advantage of highly informative categories, which determine the most part of sentence structure once correctly assigned to words.
It is demonstrated that the annotation of only pre-terminal categories is sufficient to adapt a CCG parser to new domains.
However, the solution is limited to a specific parser's architecture, making non-trivial the application of the method to the current state-of-the-art parsers~\cite{lee-lewis-zettlemoyer:2016:EMNLP2016,depccg,stanojevic-steedman-2019-ccg}, which require full parse annotation.  
Additionally, some ambiguities remain unresolved with mere supertags, especially in languages other than English (as discussed in \citet{depccg}), to which the method is not portable.  

\begin{figure*}[t]
    \centering
    \includegraphics[scale=0.60]{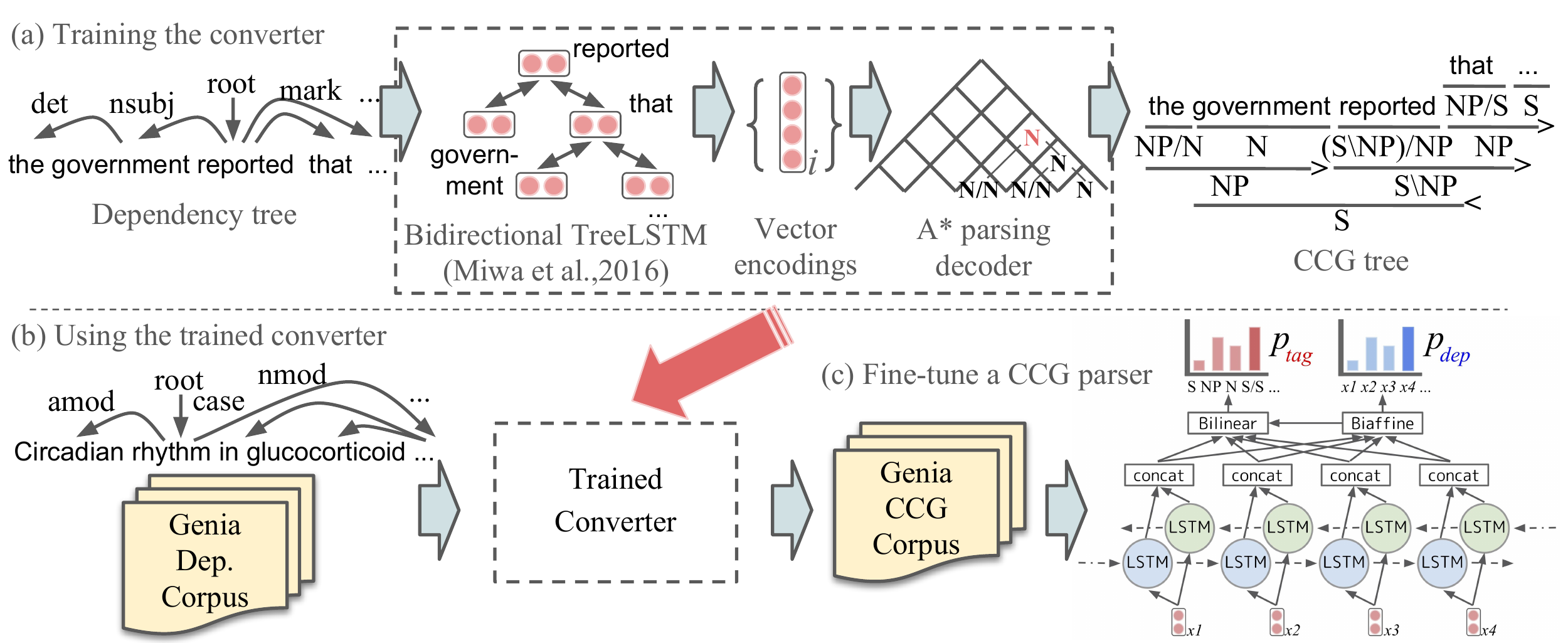}
    \caption{Overview of the proposed method. (a) A neural network-based model is trained to convert a dependency tree to a CCG one using aligned annotations on WSJ part of the Penn Treebank and the English CCGbank. (b) The trained converter is applied to an existing dependency corpus (e.g., the Genia corpus) to generate a CCGbank, (c) which is then used to fine-tune the parameters of an off-the-shelf CCG parser.}
    \label{overview}
\end{figure*}

Distributional embeddings are proven to be powerful tools for solving the issue of domain adaption, with their unlimited applications in NLP, not to mention syntactic parsing~\cite{Q14-1026,W15-2610,Peters:2018}.
Among others, \citet{P18-1110} reports huge performance boosts in constituency parsing using contextualized word embeddings~\cite{Peters:2018},
which is orthogonal to our work, and the combination shows huge gains.
Including \citet{P18-1110}, there are studies to learn from partially annotated trees \cite{Mirroshandel:2011:ALD:2206329.2206346,P16-1033,P18-1110},
again, most of which exploit specific parser architecture.

In this work, we propose a conceptually simpler approach to the issue, which is agnostic on any parser architecture, namely, {\it automatic generation of CCGbanks} (i.e., CCG treebanks)\footnote{
    In this paper, we call a treebank based on CCG grammar {\it a CCGbank}, and refer to the specific one constructed in \citet{ccgbank} as {\it the English CCGbank}.
} for new domains, by exploiting cheaper resources of dependency trees.
Specifically, we train a deep conversion model to map a dependency tree to a CCG tree, on aligned annotations of the Penn Treebank~\cite{J93-2004} and the English CCGbank~\cite{ccgbank}~(Figure~\ref{overview}a). 
When we need a CCG parser tailored for a new domain,
the trained converter is applied to a dependency corpus in that domain to obtain a new CCGbank~(\ref{overview}b), which is then used to fine-tune an off-the-shelf CCG parser~(\ref{overview}c).
The assumption that we have a dependency corpus in that target domain is not demanding given the abundance of existing dependency resources along with its developed annotation procedure, e.g., Universal Dependencies (UD) project~\cite{Nivre2016UniversalDV}, and the cheaper cost to train an annotator.

One of the biggest bottlenecks of syntactic parsing is handling of countless {\it unknown words}.
It is also true that there exist such unfamiliar input data types to our converter, e.g., disfluencies in speech and symbols in math problems.
We address these issues by {\it constrained decoding}~(\S \ref{constrained_decoding}), enabled by incorporating a parsing technique into our converter.
Nevertheless, syntactic structures exhibit less variance across textual domains than words do; our proposed converter suffers less from such unseen events, and expectedly produces high-quality CCGbanks.  

The work closest to ours is \citet{P18-1252}, where a conversion model is trained to map dependency treebanks of different annotation principles, which is used to increase the amount of labeled data in the target-side treebank.
Our work extends theirs and solves a more challenging task; the mapping to learn is to more complex CCG trees, and it is applied to datasets coming from plainly different natures (i.e., domains).
Some prior studies design conversion algorithms to induce CCGbanks for languages other than English from dependency treebanks~\cite{johan-italian,ambati-etal-2013-using}.
Though the methods may be applied to our problem, they usually cannot cover the entire dataset, consequently discarding sentences with characteristic features.
On top of that, unavoidable information gaps between the two syntactic formalisms may at most be addressed probabilistically.

To verify the generalizability of our approach, on top of the existing benchmarks on (1) {\bf biomedical texts} and (2) {\bf question sentences}~\cite{D08-1050}, we conduct parsing experiments on (3) {\bf speech conversation texts}, which exhibit other challenges such as handling informal expressions and lengthy sentences. We create a CCG version of the Switchboard corpus~\cite{switchboard}, consisting of full train/dev/test sets of automatically generated trees
and manually annotated 100 sentences for a detailed evaluation.
Additionally, we manually construct experimental data for parsing (4) {\bf math problems}~\cite{D15-1171},
for which the importance of domain adaptation is previously demonstrated \cite{P18-1110}.
We observe huge additive gains in the performance of the {\tt depccg} parser~\cite{depccg}, by combining contextualized word embeddings~\cite{Peters:2018} and our domain adaptation method:
in terms of unlabeled F1 scores, 90.68\% to 95.63\% on speech conversation, and 88.49\% to 95.83\% on math problems, respectively.\footnote{
    All the programs and resources used in this work are available at: \url{https://github.com/masashi-y/depccg}.
}
\begin{figure*}[t]
    \centering
\setpremisesend{0pt}
\setpremisesspace{1pt}
\setnamespace{0pt}
\scalebox{0.7}{
    \inference{
        \inference{
            \inference{\text{cats}}{\mathrm{N}_{\tt cats}}[]
            }{\mathrm{NP}_{\tt cats}}[un]
        \inference{
            \inference{\text{that}}{(\mathrm{NP}_x\backslash \mathrm{NP}_x)/(\mathrm{S}/\mathrm{NP}_x)}[]
            \inference{
                \inference{
                    \inference{\text{Kyle}}{\mathrm{NP}_{\tt kyle}}[]
                }{\mathrm{S}_y/(\mathrm{S}_y\backslash \mathrm{NP}_{\tt kyle})}[T]
                \inference{
                    \inference{\text{wants}}{(\mathrm{S}_{\text{wants}}\backslash \mathrm{NP}_{z,1})/(\mathrm{S}_{w,2}\backslash \mathrm{NP}_z)}[]
                    \inference{
                        \inference{\text{to}}{(\mathrm{S}_u\backslash \mathrm{NP}_v)/(\mathrm{S}_u\backslash \mathrm{NP}_v)}[]
                        \inference{\text{see}}{(\mathrm{S}_{\text{see}}\backslash \mathrm{NP}_{s,1})/\mathrm{NP}_{t,2}}[]
                        }{(\mathrm{S}_\text{see}\backslash \mathrm{NP}_v)/\mathrm{NP}_t: u = \text{see}, v = s}[$>$B]
                }{(\mathrm{S}_{\text{wants}}\backslash \mathrm{NP}_z)/\mathrm{NP}_t: w = \text{see}, z = v}[$>$B]
            }{\mathrm{S}_y/ \mathrm{NP}_t: y=\text{wants}, z = {\tt kyle}}[$>$B]
        }{\mathrm{NP}_x\backslash \mathrm{NP}_x: x = t}[$>$]
    }{\mathrm{NP}_{\text{cats}}: x = {\tt cats}}[$<$]
}
    \caption{Example CCG derivation tree for phrase \textit{cats that Kyle wants to see}.
    Categories are combined using rules such as an application rule (marked with ``$>$'', $\mathrm{X}/\mathrm{Y}~~\mathrm{Y} \Rightarrow \mathrm{X}$) and a composition rule (``$>$B'': $\mathrm{X}/\mathrm{Y}~~\mathrm{Y}/\mathrm{Z} \Rightarrow \mathrm{X}/\mathrm{Z}$).
    See \citet{steedman:syntactic_process} for the detail.
    }
    \label{example_ccg_tree}
\end{figure*}

\section{Combinatory Categorial Grammar}
\label{sec:ccg}
CCG is a lexicalized grammatical formalism, where words and phrases are assigned categories with complex internal structures.
A category $\mathrm{X}/\mathrm{Y}$ (or $\mathrm{X}\backslash \mathrm{Y}$) represents a phrase that combines with a $\mathrm{Y}$ phrase on its right (or left), and becomes an $\mathrm{X}$ phrase.
As such, a category ${(\mathrm{S}\backslash \mathrm{NP})/\mathrm{NP}}$ represents an English transitive verb which takes $\mathrm{NP}$s on both sides and becomes a sentence ($\mathrm{S}$).

The semantic structure of a sentence can be extracted using the functional nature of CCG categories.
Figure~\ref{example_ccg_tree} shows an example CCG derivation of a phrase \textit{cats that Kyle wants to see}, where categories are marked with variables and constants (e.g., ${\tt kyle}$ in $\mathrm{NP}_{\tt kyle}$), and argument ids in the case of verbs (subscripts in $(\mathrm{S}_{\text{see}}\backslash \mathrm{NP}_{s,1})/\mathrm{NP}_{t,2}$).
Unification is performed on these variables and constants in the course of derivation,
resulting in chains of equations $s = v = z = {\tt kyle}$, and $t = x = {\tt cats}$, successfully recovering the first and second argument of {\it see}: {\it Kyle} and {\it cats} (i.e., capturing {\it long-range dependencies}).
What is demonstrated here is performed in the standard evaluation of CCG parsing, where the number of such correctly predicted predicate-argument relations is calculated (for the detail, see \citet{Clark:2002:BDD:1073083.1073138}).
Remarkably, it is also the basis of CCG-based semantic parsing~\cite{abzianidze:2017:EMNLP2017Demos,martinezgomez-EtAl:2017:EACLlong,P17-1195}, where the above simple unification rule is replaced with more sophisticated techniques such as $\lambda$-calculus.

There are two major resources in CCG: the English CCGbank~\cite{ccgbank} for news texts, and the Groningen Meaning Bank~\cite{Bos2017} for wider domains, including Aesop's fables.
However, when one wants a CCG parser tuned for a specific domain,
he or she faces the issue of its high annotation cost:
\begin{itemize}
    \item The annotation requires linguistic expertise, being able to keep track of semantic composition performed during a derivation.
    \item An annotated tree must strictly conform to the grammar, e.g., inconsistencies such as combining $\mathrm{N}$ and $\mathrm{S}\backslash\mathrm{NP}$ result in ill-formed trees and hence must be disallowed.
\end{itemize}
We relax these assumptions by using {\it dependency tree}, which is a simpler representation of the syntactic structure, i.e., it lacks information of long-range dependencies and conjunct spans of a coordination structure.
However, due to its simplicity and flexibility, it is easier to train an annotator, and there exist plenty of accessible dependency-based resources, which we exploit in this work.

\section{Dependency-to-CCG Converter}
We propose a domain adaptation method based on the automatic generation of a CCGbank out of a dependency treebank in the target domain.
This is achieved by our dependency-to-CCG converter, a neural network model consisting of a dependency tree encoder and a CCG tree decoder. 

In the encoder, higher-order interactions among dependency edges are modeled with a bidirectional TreeLSTM~\cite{P16-1105}, which is important to facilitate mapping from a dependency tree to a more complex CCG tree.
Due to the strict nature of CCG grammar, we model the output space of CCG trees explicitly\footnote{
    The strictness and the large number of categories make it still hard to leave everything to neural networks to learn.
    We trained constituency-based RSP parser~\cite{P18-1110} on the English CCGbank by disguising the trees as constituency ones, whose performance could not be evaluated since most of the output trees violated the grammar.
}; our decoder is inspired by the recent success of A* CCG parsing~\cite{lewis-steedman-2014-ccg,depccg}, where the most probable valid tree is found using A* parsing~\cite{klein:a_star}.
In the following, we describe the details of the proposed converter.

Firstly, we define a probabilistic model of the dependency-to-CCG conversion process.
According to \citet{depccg}, the structure of a CCG tree ${\bm y}$ for sentence ${\bm x} = (x_1, ..., x_N)$ is almost uniquely determined\footnote{
    The uniqueness is broken if a tree contains a unary node.
} if a sequence of the pre-terminal CCG categories (supertags) ${\bm c} = (c_1, ..., c_N)$ and a dependency structure ${\bm d} = (d_1, ..., d_N)$, where $d_i \in \{0, ..., N\}$ is an index of dependency parent of $x_i$ (0 represents a root node), are provided.
Note that the dependency structure ${\bm d}$ is generally different from an input dependency tree.\footnote{
    In this work, input dependency tree is based on Universal Dependencies~\cite{Nivre2016UniversalDV}, while dependency structure ${\bm d}$ of a CCG tree is Head First dependency tree introduced in \citet{depccg}. See \S~\ref{experiment} for the detail.
 }
While supertags are highly informative about the syntactic structure~\cite{bangalore1999supertagging}, remaining ambiguities such as attachment ambiguities need to be modeled using dependencies.
Let the input dependency tree of sentence ${\bm x}$ be  
${\bm z} = ({\bm p}, {\bm d}', {\bm \ell})$,
where $p_i$ is a part-of-speech tag of $x_i$, $d'_i$ an index of its dependency parent, $\ell_i$ is the label of the corresponding dependency edge,
then the conversion process is expressed as follows:\footnote{
    Here, the independence of each $c_i$s and $d_i$s is assumed.
}
 \begin{align}
     P({\bm y} | {\bm x}, {\bm z}) & = \prod_{i = 1}^N p_{tag}(c_i | {\bm x}, {\bm z}) \prod_{i = 1}^N p_{dep}(d_i | {\bm x}, {\bm z}). \nonumber
\end{align}
Based on this formulation, we model $c_i$ and $d_i$ conditioned on a dependency tree ${\bm z}$, and search for ${\bm y}$ that maximizes $P({\bm y}| {\bm x}, {\bm z})$ using A* parsing.

\paragraph{Encoder} A bidirectional TreeLSTM consists of two distinct TreeLSTMs~\cite{P15-1150}.
A {\it bottom-up} TreeLSTM recursively computes a hidden vector ${\bm h}_i^{\uparrow}$ for each $x_i$, from vector representation ${\bm e}_i$ of the word and hidden vectors of its dependency children $\{{\bm h}_j^{\uparrow} | d'_j = i\}$.
A {\it top-down} TreeLSTM, in turn, computes ${\bm h}_i^{\downarrow}$ using ${\bm e}_i$ and a hidden vector of the dependency parent ${\bm h}_{d'_i}^{\downarrow}$.
In total, a bidirectional TreeLSTM returns concatenations of hidden vectors for all words:
${\bm h}_i = {\bm h}_i^{\uparrow} \oplus {\bm h}_i^{\downarrow}$.

We encode a dependency tree as follows,
where ${\bm e}_v$ denotes the vector representation of variable $v$, and
$\Omega$ and $\Xi_{{\bm d}'}$ are shorthand notations of the series of operations of sequential and tree bidirectional LSTMs, respectively:
\begin{align}
    {\bm e}_1, ..., {\bm e}_N & = \Omega({\bm e}_{p_1} \oplus {\bm e}_{x_1}, ..., {\bm e}_{p_N} \oplus {\bm e}_{x_N}),  \nonumber \\
    {\bm h}_1, ..., {\bm h}_N & = \Xi_{{\bm d}'}({\bm e}_1 \oplus {\bm e}_{\ell_1}, ..., {\bm e}_N \oplus {\bm e}_{\ell_N}). \nonumber
\end{align}

\paragraph{Decoder} The decoder part adopts the same architecture as in
\citet{depccg}, where $p_{dep|tag}$ probabilities are computed on top of $\{{\bm h}_i\}_{i \in [0,N]}$, using a {\it biaffine} layer~\cite{DBLP:journals/corr/DozatM16} and a bilinear layer, respectively, which are then used in A* parsing to find the most probable CCG tree. 

Firstly a biaffine layer is used to compute unigram head probabilities $p_{dep}$ as follows:
\begin{gather}
    {\bm r}_i = \psi_{child}^{dep}({\bm h}_i),~~ {\bm r}_j = \psi_{head}^{dep}({\bm h}_j), \nonumber \\
     s_{i,j} = {\bm r}_i^{\mathrm{T}} W {\bm r}_{j} + {\bm w}^{\mathrm{T}} {\bm r}_{j}, \nonumber \\
    p_{dep}(d_i = j | {\bm x}, {\bm z}) \propto \exp (s_{i,j}), \nonumber 
\end{gather}
where $\psi$ denotes a multi-layer perceptron.
The probabilities $p_{tag}$ are computed by a bilinear transformation of vector encodings $x_i$ and $x_{\hat{d}_i}$, where $\hat{d}_i$ is the most probable dependency head of $x_i$ with respect to $p_{dep}$: $\hat{d}_i = \text{arg} \max_{j} p_{dep}(d_i = j | {\bm x}, {\bm z})$.
\begin{gather}
    {\bm q}_i = \psi_{child}^{tag}({\bm h}_i),~~ {\bm q}_{\hat{d}_i} = \psi_{head}^{tag}({\bm h}_{\hat{d}_i}), \label{bilinear} \nonumber \\
    s_{i,c} = {\bm q}_i^{\mathrm{T}} W_c {\bm q}_{\hat{d}_i} +
     {\bm v}_c^{\mathrm{T}} {\bm q}_i + {\bm u}_c^{\mathrm{T}} {\bm q}_{\hat{d}_i} + b_c, \nonumber \\
    p_{tag}(c_i = c | {\bm x}, {\bm z}) \propto \exp ( s_{i,c} ). \nonumber 
\end{gather}

\paragraph{A* Parsing}
Since the probability $P({\bm y} |  {\bm x}, {\bm z})$ of a CCG tree ${\bm y}$ is simply decomposable into probabilities of subtrees,
the problem of finding the most probable tree can be solved with a chart-based algorithm.
In this work, we use one of such algorithms, A* parsing~\cite{klein:a_star}.
A* parsing is a generalization of A* search for shortest path problem on a graph, and it controls subtrees (corresponding to a node in a graph case) to visit next using a priority queue.
We follow \citet{depccg} exactly in formulating our A* parsing, and adopt an admissible heuristic by taking the sum of the max $p_{tag|dep}$ probabilities outside a subtree.
The advantage of employing an A* parsing-based decoder is 
not limited to the optimality guarantee of the decoded tree;
it enables constrained decoding, which is described next.

\section{Constrained Decoding}
\label{constrained_decoding}
While our method is a fully automated treebank generation method, there are often cases where we want to control the form of output trees by using external language resources.
For example, when generating a CCGbank for biomedical domain, it will be convenient if a disease dictionary is utilized to ensure that a complex disease name in a text is always assigned the category $\mathrm{NP}$.
In our decoder based on A* parsing, it is possible to perform such a controlled generation of a CCG tree by imposing {\it constraints} on the space of trees.
 
A constraint is a triplet $(c, i, j)$ representing a constituent of category $c$ spanning over words $x_i, ..., x_j$.
The constrained decoding is achieved by refusing to add a subtree (denoted as $(c', k, l)$, likewise, with its category and span) to the priority queue when it meets one of the conditions:
\begin{itemize}
  \setlength{\itemsep}{0cm}
    \item The spans overlap: $i < k \leq j < l$ or $k < i \leq l < j$.
    \item The spans are identical ($i = k$ and $j = l$), while the categories are different ($c \neq c'$) and no category $c''$ exists such that $c' \Rightarrow c''$ is a valid unary rule.
\end{itemize}
The last condition on unary rule is necessary to prevent structures such as $(\mathrm{NP}~(\mathrm{N}~\text{dog}))$ from being accidentally discarded, when using a constraint to make a noun phrase to be $\mathrm{NP}$.
A set of multiple constraints are imposed by checking the above conditions for each of the constraints when adding a new item to the priority queue.
When one wants to constrain a terminal category to be $c$, that is achieved by manipulating $p_{tag}$:
$p_{tag}(c | {\bm x}, {\bm z}) = 1$ and for all categories $c' \neq c$, $p_{tag}(c' | {\bm x}, {\bm z}) = 0$.

\section{Experiments}
\label{experiment}

\subsection{Experimental Settings}
We evaluate our method in terms of performance gain obtained by
fine-tuning an off-the-shelf CCG parser {\tt depccg}~\cite{depccg}, on a variety of CCGbanks obtained by converting existing dependency resources using the method.

In short, the method of {\tt depccg} is equivalent to omitting
the dependence on a dependency tree~${\bm z}$ from $P({\bm y}| {\bm x}, {\bm z})$ of our converter model, and running an A* parsing-based decoder on $p_{tag|dep}$ calculated on ${\bm h}_1, ..., {\bm h}_N = \Omega({\bm e}_{x_1}, ..., {\bm e}_{x_N})$, as in our method.
In the plain {\tt depccg}, the word representation ${\bm e}_{x_i}$ is
a concatenation of GloVe\footnote{
     \url{https://nlp.stanford.edu/projects/glove/} } vectors
     and vector representations of affixes.
As in the previous work, the parser is trained on both the English CCGbank~\cite{ccgbank} and the tri-training dataset by~\citet{depccg}.
 In this work, on top of that, we include as a baseline a setting where the affix vectors are replaced by contextualized word representation (ELMo; \citet{Peters:2018}) (${\bm e}_{x_i} = {\bm x}_{x_i}^{GloVe} \oplus {\bm x}_{x_i}^{ELMo}$),\footnote{
    We used the ``original'' ELMo model, with 1,024-dimensional word vector outputs (\url{https://allennlp.org/elmo}).
 }
 which we find marks the current best scores in the English CCGbank parsing~(Table~\ref{results_upperbound}).

 The evaluation is based on the standard evaluation metric, where the number of correctly predicted predicate argument relations is calculated~(\S\ref{sec:ccg}), where {\it labeled} metrics take into account the category through which the dependency is constructed, while {\it unlabeled} ones do not.

\paragraph{Implementation Details}
The input word representations to the converter are the concatenation of GloVe and ELMo representations.  
Each of ${\bm e}_{p_i}$ and ${\bm e}_{\ell_i}$ is randomly initialized 50-dimensional vectors, and the two-layer sequential LSTMs $\Omega$ outputs 300 dimensional vectors, as well as bidirectional TreeLSTM $\Xi_{{\bm d}'}$, whose outputs are then fed into 1-layer 100-dimensional MLPs with ELU non-linearity~\cite{DBLP:journals/corr/ClevertUH15}.
The training is done by minimizing the sum of negative log likelihood of $p_{tag|dep}$ using the Adam optimizer (with $\beta_1 = \beta_2 = 0.9$), on a dataset detailed below.

\paragraph{Data Processing}
In this work, the input tree to the converter follows Universal Dependencies (UD) v1~\cite{Nivre2016UniversalDV}.
Constituency-based treebanks are converted using the Stanford Converter\footnote{
     \url{https://nlp.stanford.edu/software/stanford-dependencies.shtml}.
     We used the version 3.9.1.} to obtain UD trees.
The output dependency structure ${\bm d}$ follows Head First dependency tree~\cite{depccg}, where a dependency arc is always from left to right.
The conversion model is trained to map UD trees in the Wall Street Journal (WSJ) portion 2-21 of the Penn Treebank~\cite{J93-2004} to its corresponding CCG trees in the English CCGbank~\cite{ccgbank}.

\begin{table}[t]
    \centering
    \begin{tabular}{lcc} \hline
         \multicolumn{1}{c}{Method}  & UF1 &  LF1 \\ \hline
        {\tt depccg} &  94.0 & 88.8 \\
        + ELMo & 94.98 & 90.51 \\ \hline \hline
        Converter &  96.48 & 92.68 \\ \hline
    \end{tabular}
    \caption{
    The performance of baseline CCG parsers and the proposed converter on WSJ23, where UF1 and LF1 represents unlabeled and labeled F1, respectively.
    }
    \label{results_upperbound}
\end{table}

\paragraph{Fine-tuning the CCG Parser}
In each of the following domain adaptation experiments, newly obtained CCGbanks are used to fine-tune the parameters of the baseline parser described above, by re-training it on the mixture of labeled examples from the new target-domain CCGbank, the English CCGbank, and the tri-training dataset.

\subsection{Evaluating Converter's Performance}
First, we examine whether the trained converter can produce high-quality CCG trees, by applying it to dependency trees in the test portion (WSJ23) of Penn Treebank and then calculating the standard evaluation metrics between the resulting trees and the corresponding gold trees (Table~\ref{results_upperbound}).
This can be regarded as evaluating the upper bound of the conversion quality, since the evaluated data comes from the same domain as the converter's training data.
Our converter shows much higher scores compared to the current best-performing {\tt depccg} combined with ELMo (1.5\% and 2.17\% up in unlabeled\allowbreak/\allowbreak labeled F1 scores), suggesting that, using the proposed converter, we can obtain CCGbanks of high quality.

\begin{table}[t]
    \centering
    \scalebox{0.84}{
    \begin{tabular}{cccc} \hline
    Relation & Parser & Converter & \# \\ \hline
        \multicolumn{4}{l}{(a)~~{\it PPs attaching to NP / VP}} \\
        $(NP\backslash \underline{\bf NP})/NP$ & 90.62 & 97.46 & 2,561 \\
        $(S\backslash NP)\backslash \underline{\bf (S\backslash NP)})/NP$ & 81.15 & 88.63 & 1,074 \\ \hdashline
    \multicolumn{4}{l}{(b)~~{\it Subject / object relative clauses}} \\
        $(NP\backslash \underline{\bf NP})/(S_{dcl}\backslash NP)$ & 93.44 & 98.71 & 307 \\
        $(NP\backslash \underline{\bf NP})/(S_{dcl}/ NP)$ & 90.48 & 93.02 & 20 \\ \hline
    \end{tabular}
    }
    \caption{
        Per-relation F1 scores of the proposed converter and {\tt depccg} + ELMo (Parser).
        ``\#'' column shows the number of occurrence of the phenomenon.
    }
    \label{results_category_breakdown}
\end{table}

Inspecting the details, the improvement is observed across the board  (Table~\ref{results_category_breakdown}); 
the converter precisely handles PP-attachment (\ref{results_category_breakdown}a), notoriously hard parsing problem, by utilizing input's {\tt pobj} dependency edges, as well as relative clauses (\ref{results_category_breakdown}b), one of well-known sources of long-range dependencies, for which the converter has to learn from the non-local combinations of edges, their labels and part-of-speech tags surrounding the phenomenon.


\begin{table}[t]
    \centering
    \begin{tabular}{lccc} \hline
         \multicolumn{1}{c}{Method}  & P & R & F1 \\ \hline
        {\tt C\&C} & 77.8 & 71.4 & 74.5 \\ 
        {\tt EasySRL} & 81.8 & 82.6 & 82.2 \\ \hdashline
        {\tt depccg} & 83.11 & 82.63 & 82.87 \\
        + ELMo & 85.87 & 85.34 & 85.61 \\
        + {\tt GENIA1000} & 85.45 & 84.49 & 84.97 \\
        + Proposed & {\bf 86.90} & {\bf 86.14} & {\bf 86.52} \\ \hline
    \end{tabular}
    \caption{Results on the biomedical domain dataset (\S\ref{sec:biomedical}).
    P and R represent precision and recall, respectively.
    The scores of {\tt C\&C} and {\tt EasySRL} fine-tuned on the {\tt GENIA1000} is included for comparison (excerpted from \citet{lewis-lee-zettlemoyer:2016:N16-1}).
    }
    \label{results_bioinfer}
\end{table}

\begin{table}[t]
    \centering
    \begin{tabular}{lccc} \hline
         \multicolumn{1}{c}{Method} & P & R & F1 \\ \hline
        {\tt C\&C} & - & - & 86.8 \\ 
        {\tt EasySRL} & 88.2 & 87.9 & 88.0 \\ \hdashline
        {\tt depccg} & 90.42 & {\bf 90.15} & {\bf 90.29} \\
        + ELMo & {\bf 90.55} & 89.86 & 90.21 \\
        + Proposed & 90.27 & 89.97 & 90.12 \\ \hline
    \end{tabular}
    \caption{Results on question sentences (\S\ref{sec:questions}).
    All of baseline {\tt C\&C}, {\tt EasySRL} and {\tt depccg} parsers are retrained on {\tt Questions} data.
    }
    \label{results_questions}
\end{table}
\subsection{Biomedical Domain and Questions}
Previous work \cite{D08-1050} provides CCG parsing benchmark datasets in biomedical texts and question sentences, each representing two contrasting challenges for a newswire-trained parser, i.e., a large amount of out-of-vocabulary words (biomedical texts), and rare or even unseen grammatical constructions (questions).

Since the work also provides small training datasets for each domain, we utilize them as well: {\tt GENIA1000} with 1,000 sentences and {\tt Questions} with 1,328 sentences, both annotated with pre-terminal CCG categories.
Since pre-terminal categories are not sufficient to train {\tt depccg}, we automatically annotate Head First dependencies using RBG parser~\cite{P14-1130}, trained to produce this type of trees (We follow \citet{depccg}'s tri-training setup).

Following the previous work, the evaluation is based on the Stanford grammatical relations~(GR; \citet{L06-1260}), a deep syntactic representation that can be recovered from a CCG tree.\footnote{
    We used their public script (\url{https://www.cl.cam.ac.uk/~sc609/candc-1.00.html}).
} 

\paragraph{Biomedical Domain}
\label{sec:biomedical}
By converting the Genia corpus~\cite{I05-2038}, we obtain a new CCGbank of 4,432 sentences from biomedical papers annotated with CCG trees.
During the process, we have successfully assigned the category $\mathrm{NP}$ to all the occurrences of complex biomedical terms by imposing constraints (\S\ref{constrained_decoding}) that $\mathrm{NP}$ spans in the original corpus be assigned the category $\mathrm{NP}$ in the resulting CCG trees as well.

Table~\ref{results_bioinfer} shows the results of the parsing experiment, where the scores of previous work ({\tt C\&C}~\cite{J07-4004} and {\tt EasySRL}~\cite{lewis-lee-zettlemoyer:2016:N16-1}) are included for reference.
The plain {\tt depccg} already achieves higher scores than these methods, and boosts when combined with ELMo (improvement of 2.73 points in terms of F1). Fine-tuning the parser on {\tt GENIA1000} results in a mixed result, with slightly lower scores. This is presumably because the automatically annotated Head First dependencies are not accurate.
Finally, by fine-tuning on the Genia CCGbank, we observe another improvement, resulting in the highest 86.52 F1 score.
 
\paragraph{Questions}
\label{sec:questions}
In this experiment, we obtain a CCG version of the QuestionBank~\cite{P06-1063},
consisting of 3,622 question sentences, excluding ones contained in the evaluation data.

Table~\ref{results_questions} compares the performance of {\tt depccg} fine-tuned on the QuestionBank, along with other baselines.
Contrary to our expectation, the plain {\tt depccg} retrained on {\tt Questions} data performs the best, with neither ELMo nor the proposed method taking any effect.
We hypothesize that, since the evaluation set contains sentences with similar constructions, the contributions of the latter two methods are less observable on top of {\tt Questions} data.
Inspection of the output trees reveals that this is actually the case; the majority of differences among parser's configurations are irrelevant to question constructions, suggesting that the models capture well the syntax of question in the data.\footnote{
    Due to many-to-many nature of mapping to GRs, the evaluation set contains relations not recoverable from the gold supertags using the provided script;
    for example, we find that from the annotated supertags of sentence \textit{How many battles did she win ?}, the $(\text{\tt amod}~\text{\tt battle}~\text{\tt many})$ relation is obtained instead of the gold $\text{\tt det}$ relation.
    This implies one of the difficulties to obtain further improvement on this set.
}

%

\subsection{Speech Conversation}
\label{sec:switchboard}
\paragraph{Setup}
We apply the proposed method to a new domain, transcription texts of speech conversation, with new applications of CCG parsing in mind.
We create the CCG version of the Switchboard corpus~\cite{switchboard}, by which, as far as we are aware of, we conduct the first CCG parsing experiments on speech conversation.\footnote{
    Since the annotated part-of-speech tags are noisy,
    we automatically reannotate them using  the {\tt core\_web\_sm} model of spaCy (\url{https://spacy.io/}), version 2.0.16.
}
We obtain a new CCGbank of 59,029\allowbreak/\allowbreak3,799\allowbreak/\allowbreak7,681 sentences for each of the train\allowbreak/\allowbreak test\allowbreak/\allowbreak development set, where the data split follows prior work on dependency parsing on this dataset~\cite{Q14-1011}.

In the conversion, we have to handle one of the characteristics of speech transcription texts, {\it disfluencies}.
In real application, it is ideal to remove disfluencies such as interjection and repairs (e.g., {\it I want a flight \underline{to Boston} \underline{um} to Denver}), prior to performing CCG-based semantic composition.
Since this corpus contains a layer of annotation that labels their occurrences, we perform constrained decoding to mark the gold disfluencies in a tree with a dummy category $\mathrm{X}$, which can combine with any category from both sides (i.e., for all category $C$, $C~\mathrm{X} \Rightarrow C$ and $\mathrm{X}~C \Rightarrow C$ are allowed).
In this work, we perform parsing experiments on texts that are clean of disfluencies, by removing $\mathrm{X}$-marked words from sentences (i.e., a pipeline system setting with an oracle disfluency detection preprocessor).\footnote{
    We regard developing joint disfluency detection and syntactic parsing method based on CCG as future work.
}

\begin{table}[t]
    \centering
    \begin{tabular}{cp{6.4cm}} \hline
    a. & {\small\it we should cause it does help} \\
    b. & {\small\it the only problem i see with term limitations is that i think that the bureaucracy in our government as is with most governments is just so complex that there is a learning curve and that you ca n't just send someone off to washington and expect his first day to be an effective congress precision}\\ \hline
    \end{tabular}
\caption{Example sentences from the manually annotated subset of Switchboard test set.}
\label{switchboard_example_sentences}
\end{table}

\begin{table}[t]
    \centering
    \begin{tabular}{lc} \hline
        \multicolumn{1}{c}{Error type} & \# \\ \hline
        PP-attachment & 3 \\
        Adverbs attaching wrong place & 11 \\
        Predicate-argument & 5 \\
        Imperative & 2 \\
        Informal functional words & 2 \\
        Others & 11 \\ \hline
    \end{tabular}
    \caption{Error types observed in the manually annotated Switchboard subset data.}
    \label{results_switchboard_breakdown}
\end{table}

\begin{table}[t]
    \centering
    \scalebox{0.88}{
    \begin{tabular}{l|ccc|cc} \hline
        \multirow{2}{*}{Method} & \multicolumn{3}{c|}{\bf Whole} & \multicolumn{2}{c}{\bf Subset} \\
          & P & R & F1 & UF1 & LF1 \\ \hline
        {\tt depccg} & 74.73 & 73.91 & 74.32 & 90.68 & 82.46 \\
        + ELMo & 75.76 & 76.62 & 76.19 & 93.23 & 86.46 \\
        + Proposed & {\bf 78.03} & {\bf 77.06} & {\bf 77.54} & {\bf 95.63} & {\bf 92.65} \\ \hline
    \end{tabular}
    }
    \caption{Results on speech conversation texts (\S\ref{sec:switchboard}), on the whole test set and the manually annotated subset.}
    \label{results_switchboard}
\end{table}
    
\begin{figure*}[t]
    \centering
\setpremisesend{0pt}
\setpremisesspace{1pt}
\setnamespace{0pt}
\scalebox{0.67}{
\inference{
    \inference{
        \inference{\text{if}}{((\mathrm{S}\backslash \mathrm{NP})/(\mathrm{S}\backslash \mathrm{NP}))/\mathrm{S}_{dcl}}[]
        & \inference{
            \inference{
                \inference{
                    \inference{\text{CD}}{\mathrm{N}}[]
                    }{\mathrm{NP}}[un]
                & \inference{
                    \inference{\text{=}}{(\mathrm{S}_{dcl}\backslash \mathrm{NP})/\mathrm{NP}}[]
                    & \inference{
                        \inference{\text{8}}{\mathrm{N}}[]
                        }{\mathrm{NP}}[un]
                    }{\mathrm{S}_{dcl}\backslash \mathrm{NP}}[$>$]
                }{\mathrm{S}_{dcl}}[$<$]
            & \inference{
                \inference{\text{and}}{\mathrm{conj}}[]
                \inference{
                    \inference{
                        \inference{\text{BE}}{\mathrm{N}}[]
                        }{\mathrm{NP}}[un]
                    & \inference{
                        \inference{\text{=}}{(\mathrm{S}_{dcl}\backslash \mathrm{NP})/\mathrm{NP}}[]
                        & \inference{
                            \inference{\text{2}}{\mathrm{N}}[]
                            }{\mathrm{NP}}[un]
                        }{\mathrm{S}_{dcl}\backslash \mathrm{NP}}[$>$]
                    }{\mathrm{S}_{dcl}}[$<$]
                }{\mathrm{S}_{dcl}\backslash \mathrm{S}_{dcl}}[$\Phi$]
            }{\mathrm{S}_{dcl}}[$<$]
        }{(\mathrm{S}\backslash \mathrm{NP})/(\mathrm{S}\backslash \mathrm{NP})}[$>$]
    & \inference{
        \inference{
            \inference{\text{,}}{\mathrm{,}}[]
            & \inference{
                \inference{\text{find}}{(\mathrm{S}_{dcl}\backslash \mathrm{NP})/\mathrm{NP}}[]
                & \inference{
                    \inference{\text{AE}}{\mathrm{N}}[]
                    }{\mathrm{NP}}[un]
                }{\mathrm{S}_{dcl}\backslash \mathrm{NP}}[$>$]
            }{\mathrm{S}_{dcl}\backslash \mathrm{NP}}[rp]
        & \inference{\text{.}}{\mathrm{.}}[]
        }{\mathrm{S}_{dcl}\backslash \mathrm{NP}}[rp]
    }{\mathrm{S}_{dcl}\backslash \mathrm{NP}}[$>$]
}
    \caption{Parse output by the re-trained parser
    for sentence \textit{if CD = 8 and BE = 2, find AE.} from math problems.
    }
    \label{parse_result}
\end{figure*}

Another issue in conducting experiments on this dataset is evaluation.
Since there exists no evaluation protocol for CCG parsing on speech texts,
we evaluate the quality of output trees by two procedures; in the first experiment, we parse the entire test set, and convert them to constituency trees using a method by \citet{P12-2021}.\footnote{
    \url{https://github.com/jkkummerfeld/berkeley-ccg2pst}
} 
We report labeled bracket F1 scores between the resulting trees and the gold trees in the true Switchboard corpus, using the EVALB script.\footnote{
    \url{https://nlp.cs.nyu.edu/evalb/}
}
However, the reported scores suffer from the compound effect of failures in CCG parsing as well as ones occurred in the conversion to the constituency trees.
To evaluate the parsing performance in detail, the first author manually annotated a subset of randomly sampled 100 sentences from the test set.
Sentences with less than four words are not contained, to exclude short phrases such as nodding.  
Using this test set, we report the standard CCG parsing metrics.
Sentences from this domain exhibit other challenging aspects (Table~\ref{switchboard_example_sentences}), such as less formal expressions (e.g., use of {\it cause} instead of {\it because}) (\ref{switchboard_example_sentences}a), and lengthy sentences with many embedded phrases (\ref{switchboard_example_sentences}b).\footnote{
    Following \citet{Q14-1011}, sentences in this data are fully lower-cased and contain no punctuation.
}

\begin{table}[t]
    \centering
    \begin{tabular}{lcccccc} \hline
         \multicolumn{1}{c}{Method}  & UF1 & LF1 \\ \hline
        {\tt depccg} & 88.49 & 66.15 \\
        + ELMo & 89.32 & 70.74 \\
        + Proposed & {\bf 95.83} & {\bf 80.53} \\ \hline
    \end{tabular}
    \caption{Results on math problems (\S\ref{sec:math}).}
    \label{results_math_problem}
\end{table}

\paragraph{Results} 
On the whole test set, {\tt depccg} shows consistent improvements when combined with ELMo and the proposed method, in the constituency-based metrics ({\bf Whole} columns in Table~\ref{results_switchboard}).
Though the entire scores are relatively lower, the result suggests that the proposed method is effective to this domain on the whole.  
By directly evaluating the parser's performance in terms of predicate argument relations ({\bf Subset} columns), we observe that it actually recovers the most of the dependencies, with the fine-tuned {\tt depccg} achieving as high as 95.63\% unlabeled F1 score.

We further investigate error cases of the fine-tuned {\tt depccg} in the subset dataset (Table~\ref{results_switchboard_breakdown}).
The tendency of error types is in accordance with other domains, with frequent errors in PP-attachment and predicate-argument structure, and seemingly more cases of attachment errors of adverbial phrases (11 cases), which occur in lengthy sentences such as in Table~\ref{switchboard_example_sentences}b.
Other types of error are failures to recognize that the sentence is in imperative form (2 cases), and ones in handling informal functional words such as {\it cause} (Table~\ref{switchboard_example_sentences}a).
We conclude that the performance on this domain is as high as it is usable in application. Since the remaining errors are general ones, they will be solved by improving general parsing techniques.

\subsection{Math Problems}
\label{sec:math}
\paragraph{Setup} Finally, we conduct another experiment on parsing math problems.
Following previous work of constituency parsing on math problem~\cite{P18-1110}, we use the same train/test sets by \citet{D15-1171},  consisting of 63/62 sentences respectively, and see if a CCG parser can be adapted with the small training samples. 
Again, the first author annotated both train/test sets, dependency trees on the train set, and CCG trees on the test set, respectively.
In the annotation, we follow the manuals of the English CCGbank and the UD.
We regard as an important future work extending the annotation to include fine-grained feature values in categories, e.g., marking a distinction between integers and real numbers~\cite{P17-1195}.
Figure~\ref{parse_result} shows an example CCG tree from this domain, successfully parsed by fine-tuned {\tt depccg}.

\paragraph{Results} Table~\ref{results_math_problem} shows the F1 scores of {\tt depccg} in the respective settings.
Remarkably, we observe huge additive performance improvement.
While, in terms of labeled F1, ELMo contributes about 4 points on top of the plain {\tt depccg}, adding the new training set (converted from dependency trees) improves more than 10 points.\footnote{
    Note that, while in the experiment on this dataset in the previous constituency parsing work~\cite{P18-1110}, they evaluate on partially annotated (unlabeled) trees, we perform the ``full'' CCG parsing evaluation, employing the standard evaluation metrics.
    Given that, the improvement is even more significant.
}
Examining the resulting trees, we observe that the huge gain is primarily involved with expressions unique to math.
Figure~\ref{parse_result} is one of such cases, which the plain {\tt depccg} falsely analyzes as one huge $\mathrm{NP}$ phrase.
However, after fine-tuning, it successfully produces the correct ``If $\mathrm{S}_1$ and $\mathrm{S}_2$, $\mathrm{S}_3$'' structure, recognizing that the equal sign is a predicate.

\section{Conclusion}
In this work, we have proposed a domain adaptation method for CCG parsing, based on the automatic generation of new CCG treebanks from dependency resources.  
We have conducted experiments to verify the effectiveness of the proposed method on diverse domains: on top of existing benchmarks on biomedical texts and question sentences,
we newly conduct parsing experiments on speech conversation and math problems.
Remarkably, when applied to our domain adaptation method, the improvements in the latter two domains are significant, with the achievement of more than 5 points in the unlabeled metric.
%

\section*{Acknowledgments}
We thank the three anonymous reviewers for their insightful comments.
This work was in part supported by JSPS KAKENHI
Grant Number JP18J12945, and also by JST AIP-PRISM Grant Number JPMJCR18Y1, Japan.

\bibliography{acl2019}
\bibliographystyle{acl_natbib}


\end{document}